\documentclass[runningheads]{llncs}

\usepackage{amssymb}
\usepackage{graphicx}
\usepackage{amsmath}
\usepackage{pgfplots}
\usepgfplotslibrary{groupplots}
\usepackage{array,multirow}

\usepackage{xcolor}
\usepackage{makecell}
\usepackage{hyperref}

\hypersetup{
colorlinks=false,
linkcolor=blue,
filecolor=blue,
urlcolor=blue,
citecolor=blue}

\begin{document}

\title{Dataset and Evaluation algorithm design for GOALS Challenge}
\titlerunning{Dataset and Evaluation algorithm design for GOALS Challenge}

\mainmatter
\author{Huihui Fang\inst{1} \and
Fei Li\inst{2} \and
Huazhu Fu \inst{3} \and
Junde Wu\inst{1} \and
Xiulan Zhang\inst{2}\thanks{Corresponding authors.} \and
Yanwu Xu\inst{1\star}}

\authorrunning{Fang et al.}

\institute{ Intelligent Healthcare Unit, Baidu Inc., Beijing, China\\ \email{ywxu@ieee.org} \and State Key Laboratory of Ophthalmology, Zhongshan Ophthalmic Center, Sun Yat-sen University, Guangdong Provincial Key Laboratory of Ophthalmology and Visual
Science, Guangzhou, China \\ \email{zhangxl2@mail.sysu.edu.cn}\and Institute of High Performance Computing, Agency for Science, Technology and Research, Singapore
}
\maketitle              
%
\begin{abstract}   
Glaucoma causes irreversible vision loss due to damage to the optic nerve, and there is no cure for glaucoma.OCT imaging modality is an essential technique for assessing glaucomatous damage since it aids in quantifying fundus structures. To promote the research of AI technology in the field of OCT-assisted diagnosis of glaucoma, we held a Glaucoma OCT Analysis and Layer Segmentation (GOALS) Challenge in conjunction with the International Conference on Medical Image Computing and Computer Assisted Intervention (MICCAI) 2022 to provide data and corresponding annotations for researchers studying layer segmentation from OCT images and the classification of glaucoma. This paper describes the released 300 circumpapillary OCT images, the baselines of the two sub-tasks, and the evaluation methodology. The GOALS Challenge is accessible at \url{https://aistudio.baidu.com/aistudio/competition/detail/230}. 

\end{abstract}

\section{Introduction}
Glaucoma is a chronic neurodegenerative condition that is one of the leading causes of irreversible blindness in the world. It is a multifactorial optic neuropathy characterized by progressive neurodegeneration of retinal ganglion cells (RGCs) and their axons, resulting in retinal nerve fiber layer (RNFL) attenuation, a specific pattern of damage to the optic nerve head, and visual field loss~\cite{sehi2013retinal}. In 2020, about 80 million people have glaucoma worldwide~\cite{Glaucoma_FactsandFigures-web}, and this number is projected to be 111.8 million in 2040~\cite{tham2014global}. Optical coherence tomography (OCT) is a powerful tool for diagnosing ocular diseases because of its no radiation, non-invasive, high resolution, high detection sensitivity and other characteristics~\cite{puzyeyeva2011high,yaqoob2005spectral}. In contrast to color fundus images, which can only provide information about the surface of the retina, OCT images can provide cross-sectional information about fundus structures. The retinal structures contain RNFL, ganglion cell-inner plexiform layer (GCIPL), inner nuclear layer (INL), outer plexiform layer (OPL), outer nuclear layer (ONL), external limiting membrane (ELM), inner photoreceptor segment, inner/outer photoreceptor segment junction, outer photoreceptor segment, retinal pigment epithelium (RPE) interdigitation, RPE/Bruch’s membrane complex, as well as choroid layer~\cite{mohandass2017retinal,medeiros2009detection}. In the diagnosis of glaucoma, the disease can be judged by observing changes in the thickness of the optic nerve fiber layer, etc., which is easier to detect early glaucoma than by observing fundus color images. The circumpapillary OCT image corresponds to a circular scan located around the optic nerve head, where a wealth of information about the different retinal layers can be found~\cite{garcia2020glaucoma}. 

Currently, there are only a limited number of OCT datasets~\cite{rasti2017macular,gholami2020octid} available in public, to facilitate researchers to conduct research on OCT images, we hold a GOALS Challenge in conjunction with MICCAI 2022, aiming to provide circumpapillary OCT images for studying layer segmentation and glaucoma classification. This paper mainly introduces the 300 circumpapillary OCT images released in the GOALS challenge, and provides baselines for the two sub-tasks (Layer segmentation, and Glaucoma classification). Meanwhile, the evaluation methods are described in detail.  

\section{Dataset}
The 300 circumpapillary OCT images are randomly selected from previous glaucoma study cohorts collected over the past five years in Zhongshan Ophthalmic Center, Sun Yat-sen University, Guangzhou, China. The images are all acquired by using a TOPCON DRI Swept Source OCT~\cite{DRIOCT}. The acquired images are saved in BMP format with a resolution of $1024 \times 247$ or in JPG format with $1270 \times 763$. In the GOALS Challenge, we store the images in PNG format with a resolution of $1100 \times 800$. The summary of the GOALS dataset and its population demographic are shown in Table~\ref{tab:dataset}.

\begin{table}[htbp]
  \centering
  \caption{Summary of the GOALS dataset and the demographic of the population.}
    \begin{tabular}{cccccc}
        \hline
          &       & \multicolumn{1}{c}{Person} & \multicolumn{1}{c}{Eyes} & Age   & \multicolumn{1}{c}{Gender (Female)} \\
         \hline
    \multirow{2}[0]{*}{Total dataset} & Total  & 66    & 99    & 45.91±15.04 & 36.40\% \\
          & Glaucoma & 13    & 22    & 44.59±12.77 & 30.80\% \\
    \multirow{2}[0]{*}{Training set} & Total  & 16    & 24    & 39.08±14.08 & 31.30\% \\
          & Glaucoma & 4     & 7     & 40.86±9.23 & 25\% \\
    \multirow{2}[0]{*}{Preliminary set} & Total  & 19    & 30    & 44.8±16.32 & 42.11\% \\
          & Glaucoma & 4     & 7     & 42.92±17.09 & 75\% \\
    \multirow{2}[0]{*}{Final set} & Total  & 31    & 45    & 50.05±14.48 & 35.50\% \\
          & Glaucoma & 5     & 8     & 50±8.48 & 0\% \\
          \hline
    \end{tabular}%
  \label{tab:dataset}%
\end{table}%

\begin{figure}
    \centering
    \includegraphics[width=1\linewidth]{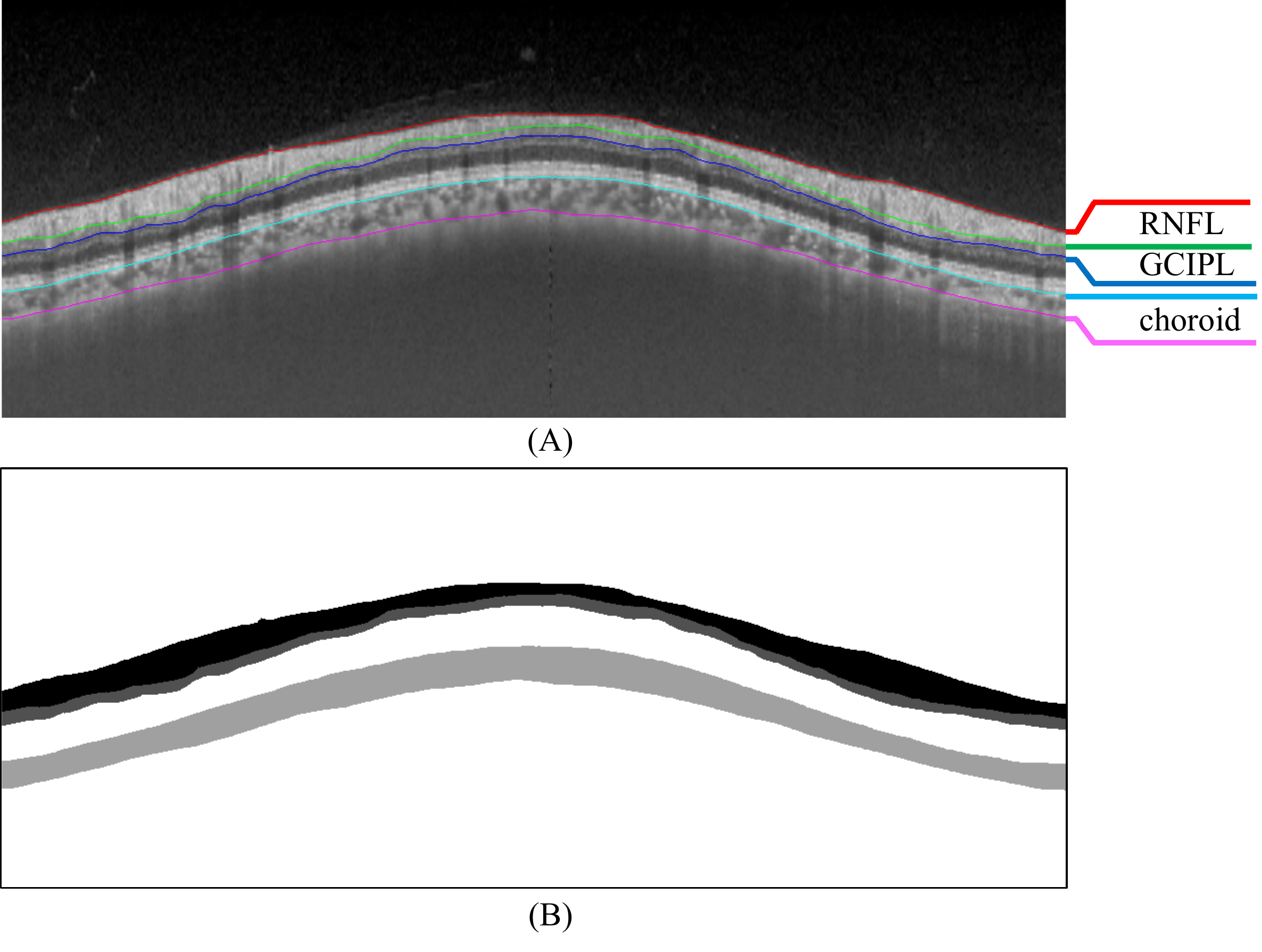}
    \caption{Schematic diagram of the annotations for RNFL, GCIPL, and choroid layer segmentation.(A) Annotations for the boundaries of the targets; (B) Segmentation masks.}
    \label{fig:annotation}
\end{figure}

The GOALS dataset provides the glaucoma labels and the segmentation masks of RNFL, GCIPL and choriod layers in the circumpapillary OCT images. The glaucoma labels are determined by the clinical records, which can reflect the findings from a series of eye examinations. The annotations of the layer segmentation are manually marked by the ophthalmologists of the iChallenge-GOALS study group. The iChallenge-GOALS study group contains ten ophthalmologists from different hospitals who have been working in the ocular field for 5 years or more. The 300 images are randomly divided into 2 subsets, and repeating this operation 5 times, we obtain 10 subsets of the data, where each image appears in 5 different subsets. The 10 subsets are randomly assigned to the 10 ophthalmologists, that is, each image is labeled by 5 different ophthalmologists. The results of the 5 initial labeling results are then aggregated by a more senior ophthalmologist for fusion. Specifically, The ophthalmologist need to delineate the upper and lower margin of the RNFL, GCIPL and choroid regions, as shown in Fig.~\ref{fig:annotation}(A).After that, the senior ophthalmologist analyze the 5 initial annotations of each image, remove the annotations with large deviations, and average the remaining initial annotations as the final annotation result for each image. We then assigne different pixel values to pixels within the boundaries of RNFL, GCIPL and choroid layer to obtain the final ground truths of the layer segmentation (RNFL: 0, GCIPL:80, choroid:160, elsewhere:255, as shown in Fig.~\ref{fig:annotation}(B)).

In GOALS Challenge, 300 images are divided into three partitions according to the patient dimension, i.e. the images acquired from the same patient's eyes are guaranteed to be divided into the same partition. These three data partitions correspond to the training set, the preliminary set, and the final set. The data in the training set contains the original OCT images and their glaucoma labels and layer segmentation masks, which are used for training models. While the preliminary and final sets only contain the original OCT images, which are used for testing models in preliminary and final rounds. 

\section{Baseline}
\begin{figure}
    \centering
    \includegraphics[width=1\linewidth]{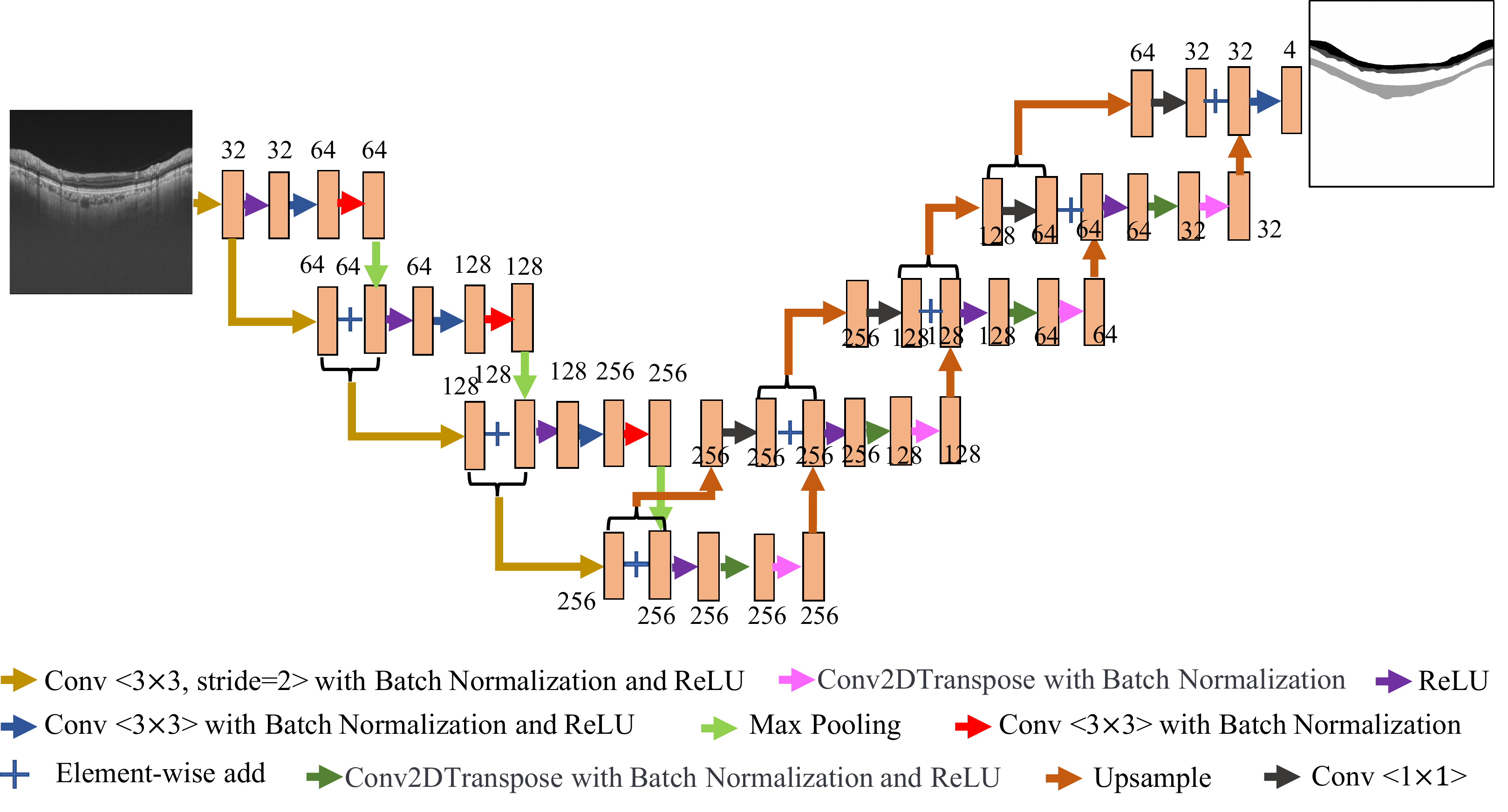}
    \caption{A baseline framework for OCT layer segmentation.}
    \label{fig:layer_seg}
\end{figure}

\begin{figure}
    \centering
    \includegraphics[width=1\linewidth]{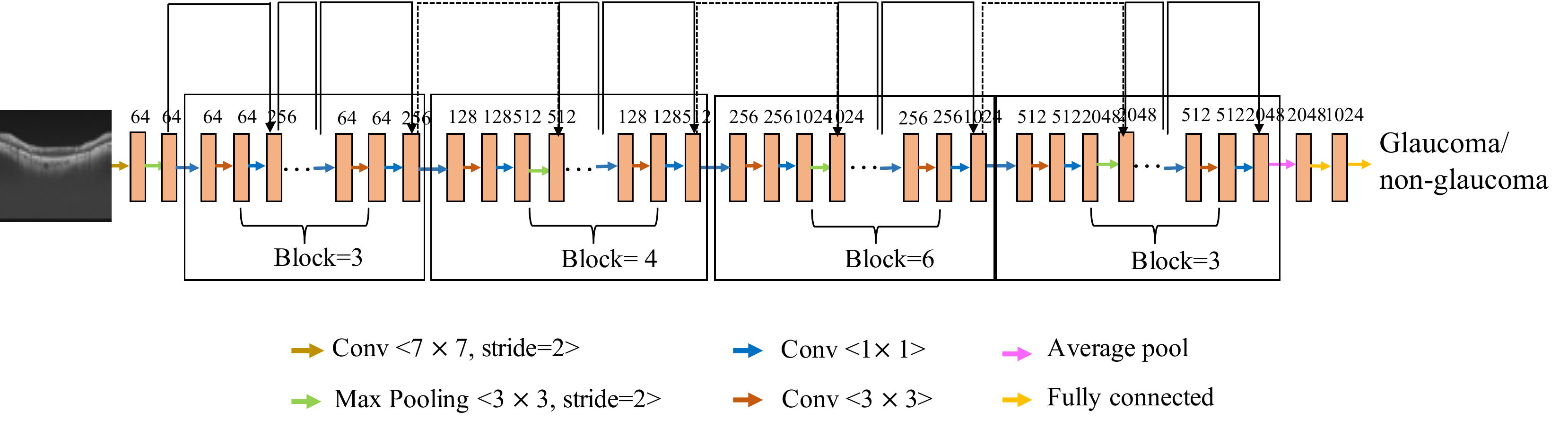}
    \caption{A baseline framework for glaucoma classification.}
    \label{fig:glaucoma_classification}
\end{figure}

We design a baseline model for each of the two challenge sub-tasks. As shown in Figs.~\ref{fig:layer_seg} and~\ref{fig:glaucoma_classification}, we utilize a U-shape network with residual concept to achieve the layer segmentation, and utilize a ResNet50 to perform the glaucoma classification. The baseline codes are avaliable at \url{https://aistudio.baidu.com/aistudio/competition/detail/230/0/related-material}. 

We implement the baselines via PaddlePaddle. During training, we use an Adam optimizer with learning rate = $10^{-3}$ in the layer segmentation task, as well as with learning rate = $10^{-6}$ in the glaucoma classification task. The training procedure consist of 3000 iterations and 1000 iterations for layer segmentation and glaucoma classification with a Nvidia Tesla V100-SXM2 GPU, respectively. The batch sizes are 8 for both tasks.

\section{Evaluation}
In this section, we introduce the evaluation metrics for the two challenge sub-tasks. For the layer segmentation task, a DICE coefficient and a mean Euclidean distance (MED) are used to evaluate the predicted region and boundary, respectively. For the glaucoma classification,  the weighted combination of sensitivity (Sen), specificity (Spe), accuracy (Acc), $F_{1}$ score, and area under the receiver operating characteristic curve (AUC) are utilized to evaluate the predicted results.

\subsection{Task 1: Layer Segmentation}
To measure the accuracy of the predicted region, we use the frequently-used DICE coefficient in the segmentation task:
\begin{equation}
    Dice = \frac{2|X\cap Y|}{|X|+|Y|}
\end{equation}
where, $X$ represents the segmented target pixel point set in the ground truth; $Y$ represents the segmented pixel point set in the prediction result; $|X\cap Y|$ represents the intersection between $X$ and $Y$; $|X|$ and $|Y|$ represent the number of the elements of $X$ and $Y$. The formula for calculating the score corresponding to the DICE metric is 
\begin{equation}
    Score_{Dice} = DICE \times 10
\end{equation}

In addition to evaluating the accuracy of the region segmentation, we also evaluate the accuracy of the boundary of the segmentation results by using Euclidean distance, due to the importance of the boundaries between the structural layers in the OCT images. Specifically, we first traverse each pixel on the predicted boundary, and calculate the Euclidean distance from each pixel to the nearest pixel on the gold standard boundary. Then the sum of the above Euclidean distances is averaged based on the number of pixels on the predicted boundary:
\begin{equation}
    MED = \frac{1}{N} \sum_{i=1}^N \sqrt{(x_{i}-x_{i}^{0})^2 + (y_{i}-y_{i}^0)^2}
\end{equation}  
where $N$ is the number of pixels on the predicted boundary, $(x_i, y_i)$ is the $i$th pixel on the predicted boundary, and $(x_i^0, y_i^0)$ is the nearest pixel on the boundary of the ground truth to $(x_i, y_i)$. The score corresponding to the MED metric is calculated by
\begin{equation}
    Score_{MED} = (MED + 1)^{-0.3}
\end{equation}

Since the layer segmentation task contains the segmentation of the three regions of RNFL, GCIPL and choroid, the DICE and MED metrics of these three regions should be taken into account in the score calculation. In addition, because the RNFL layer is more important for the diagnosis of glaucoma, we assign higher weights to the scores obtained from RNFL segmentation:
\begin{equation}
    Score_{task1} = 0.4 \times Score_{RNFL} + 0.3 \times Score_{GCIPL} + 0.3 \times Score_{choroid}
\end{equation}
\begin{equation}
\begin{split}
     Score_{region} = &0.5 \times Score_{DICE_{region}} + 0.5 \times Score_{MED_{region}}, \\
    & region \in \{RNFL, GCIPL, choroid\}
\end{split}
\end{equation}

\subsection{Task 2: Glaucoma Classification}
For glaucoma classification, we adopt five common metrics including Sen, Spe, Acc, $F_1$, and AUC:
\begin{equation}
    Sen = \frac{TP}{TP+FN}
\end{equation}
\begin{equation}
    Spe = \frac{TN}{TP+FP}
\end{equation}
\begin{equation}
    Acc = \frac{TP+TN}{TP+FN+TN+FP}
\end{equation}
\begin{equation}
    F_{1} = \frac{2\times TP}{2 \times TP+FP+FN}
\end{equation}
where $TP$, $TN$, $FP$ and $FN$ represent the numbers of true positive, true negative, false positive, and false negative detection of the glaucoma. Sen,Spe and Acc can reflect the proportions of positive samples, negative samples and all samples predicted correctly, respectively. $F_1$ provides a overall metric of the model's ability to detect comprehensively and accurately. And AUC reflects the classification ability of the model when the positive and negative samples are unbalanced. In our evaluation framework, these metrics are implemented via scikit-learn package~\cite{pedregosa2011scikit}, which is an open source machine learning toolkit base on Python.   
Since the GOALS dataset has a balanced distribution of positive and negative samples, we assign the lowest weight to the AUC metric in the score calculation.
\begin{equation}
    Score_{task2} = (0.1 \times AUC + 0.25 \times Sen + 0.25 \times Spe + 0.2 \times ACC + 0.2 \times F_{1})\times 10
\end{equation}

Based on the results of the baseline model, we find that positive and negative samples in the GOALS dataset have obvious distinguishable image features, and therefore score high in Task 2. Hence, for preliminary and final rounds, a lower weight is assigned to Task 2 in the score calculation:
\begin{equation}
\begin{split}
    Score_{round} = 0.8 \times Score_{task1} & + 0.2 \times Score_{task2},\\
    & round \in \{preliminary, final\}
\end{split}
\end{equation}
Since the preliminary leaderboard is visible to the players, one can adjust the model parameters or strategies to get the best prediction on the preliminary set. To avoid players' results from getting overfitting results on the preliminary set and getting high scores, we assign lower weights to the preliminary score when counting the total challenge scores. Hence, the total score is:
\begin{equation}
    Score = 0.3 \times Score_{preliminary} + 0.7 \times Score_{final}
\end{equation}

Based on the evaluation criteria, our baselines receive 7.2802 score on the preliminary set and 7.2398 score on the final set. The results of each specific evaluation index are shown in Table.~\ref{tab:baseline_score}.

\begin{table}[htbp]
  \centering
  \caption{The evaluation results of the baseline model on different datasets.}
    \begin{tabular}{cccc}
    \hline
    \multicolumn{2}{c}{Dataset} & \multicolumn{1}{c}{Preliminary set} & \multicolumn{1}{c}{Final set} \\
    \hline
    \multicolumn{2}{c}{Score} & \multicolumn{1}{c}{7.2802} & \multicolumn{1}{c}{7.2398} \\
    \hline
    \multirow{6}[0]{*}{Layer Segmentation} & RNFL\_DICE & 0.8161 & 0.8433 \\
          & RNFL\_ED & 4.0597 & 4.151 \\
          & GCIPL\_DICE & 0.6295 & 0.6234 \\
          & GCIPL\_ED & 3.318 & 3.6011 \\
          & choroid\_DICE & 0.8193 & 0.8746 \\
          & choroid\_ED & 8.9155 & 9.8953 \\
    \hline
    \multirow{5}[0]{*}{Glaucoma Classification} & AUC   & 0.9984 & 0.9927 \\
          & F1    & 0.9346 & 0.8829 \\
          & ACC   & 0.93  & 0.8687 \\
          & SEN   & 1     & 1 \\
          & SPE   & 0.86  & 0.74 \\
    \hline
    \end{tabular}%
  \label{tab:baseline_score}%
\end{table}%

\section{Conclusion}
In this paper, we introduce the GOALS Challenge at MICCAI 2022. In the challenge, we focus on OCT which is a powerful imaging technology for glaucoma diagnostics. We design two challenge sub-tasks, including OCT layer segmentation of RNFL, GCIPL and choroid, and glaucoma classification. The dataset collection and labeling process, as well as the result evaluation design are described in detail in the paper. GOALS Challenge dataset and evaluation framework are publicly accessible through the AI Studio website at \url{https://aistudio.baidu.com/aistudio/competition/detail/230}. Participants are welcome to join the GOALS Challenge and submit their predicted results on the website.

\bibliographystyle{unsrt}
\bibliography{mybib}

\begin{thebibliography}{10}

\bibitem{sehi2013retinal}
Mitra Sehi, Xinbo Zhang, David~S Greenfield, YunSuk Chung, Gadi Wollstein,
  Brian~A Francis, Joel~S Schuman, Rohit Varma, David Huang, Advanced~Imaging
  for Glaucoma Study~Group, et~al.
\newblock Retinal nerve fiber layer atrophy is associated with visual field
  loss over time in glaucoma suspect and glaucomatous eyes.
\newblock {\em American journal of ophthalmology}, 155(1):73--82, 2013.

\bibitem{Glaucoma_FactsandFigures-web}
Glaucoma: Facts and figures.
\newblock
  \url{https://www.brightfocus.org/glaucoma/article/glaucoma-facts-figures}.

\bibitem{tham2014global}
Yih-Chung Tham, Xiang Li, Tien~Y Wong, Harry~A Quigley, Tin Aung, and Ching-Yu
  Cheng.
\newblock Global prevalence of glaucoma and projections of glaucoma burden
  through 2040: a systematic review and meta-analysis.
\newblock {\em Ophthalmology}, 121(11):2081--2090, 2014.

\bibitem{puzyeyeva2011high}
Olena Puzyeyeva, Wai~Ching Lam, John~G Flanagan, Michael~H Brent, Robert~G
  Devenyi, Mark~S Mandelcorn, Tien Wong, and Christopher Hudson.
\newblock High-resolution optical coherence tomography retinal imaging: a case
  series illustrating potential and limitations.
\newblock {\em Journal of ophthalmology}, 2011, 2011.

\bibitem{yaqoob2005spectral}
Zahid Yaqoob, Jigang Wu, and Changhuei Yang.
\newblock Spectral domain optical coherence tomography: a better oct imaging
  strategy.
\newblock {\em Biotechniques}, 39(6):S6--S13, 2005.

\bibitem{mohandass2017retinal}
G~Mohandass, R~Ananda Natarajan, and S~Sendilvelan.
\newblock Retinal layer segmentation in pathological sd-oct images using
  boisterous obscure ratio approach and its limitation.
\newblock {\em Biomedical and Pharmacology Journal}, 10(3):1585--1591, 2017.

\bibitem{medeiros2009detection}
Felipe~A Medeiros, Linda~M Zangwill, Luciana~M Alencar, Christopher Bowd,
  Pamela~A Sample, Remo Susanna, and Robert~N Weinreb.
\newblock Detection of glaucoma progression with stratus oct retinal nerve
  fiber layer, optic nerve head, and macular thickness measurements.
\newblock {\em Investigative ophthalmology \& visual science},
  50(12):5741--5748, 2009.

\bibitem{garcia2020glaucoma}
Gabriel Garc{\'\i}a, Roc{\'\i}o del Amor, Adri{\'a}n Colomer, and Valery
  Naranjo.
\newblock Glaucoma detection from raw circumpapillary oct images using fully
  convolutional neural networks.
\newblock In {\em 2020 IEEE international conference on image processing
  (ICIP)}, pages 2526--2530. IEEE, 2020.

\bibitem{rasti2017macular}
Reza Rasti, Hossein Rabbani, Alireza Mehridehnavi, and Fedra Hajizadeh.
\newblock Macular oct classification using a multi-scale convolutional neural
  network ensemble.
\newblock {\em IEEE transactions on medical imaging}, 37(4):1024--1034, 2017.

\bibitem{gholami2020octid}
Peyman Gholami, Priyanka Roy, Mohana~Kuppuswamy Parthasarathy, and Vasudevan
  Lakshminarayanan.
\newblock Octid: Optical coherence tomography image database.
\newblock {\em Computers \& Electrical Engineering}, 81:106532, 2020.

\bibitem{DRIOCT}
Dri oct triton series.
\newblock
  \url{https://topconhealthcare.eu/uploads/media/60cb7b98ea585/topcon-triton-brochure-rev5-27-05-21-e325-lores.pdf}.

\bibitem{pedregosa2011scikit}
Fabian Pedregosa, Ga{\"e}l Varoquaux, Alexandre Gramfort, Vincent Michel,
  Bertrand Thirion, Olivier Grisel, Mathieu Blondel, Peter Prettenhofer, Ron
  Weiss, Vincent Dubourg, et~al.
\newblock Scikit-learn: Machine learning in python.
\newblock {\em the Journal of machine Learning research}, 12:2825--2830, 2011.

\end{thebibliography}

\end{document}